\documentclass[10pt,twocolumn,letterpaper]{article}

\usepackage{cvpr}
\usepackage{times}
\usepackage{epsfig}
\usepackage{graphicx}
\usepackage{amsmath}
\usepackage{amssymb}
\usepackage{multirow}

\usepackage{booktabs}
\usepackage{algorithm}
\usepackage{algorithmicx}
\usepackage{multicol,multirow}
\usepackage{subfigure}
\usepackage[dvipsnames]{xcolor}

\usepackage{pifont}
\usepackage{setspace}

\usepackage[pagebackref=true,breaklinks=true,letterpaper=true,colorlinks,bookmarks=false]{hyperref}

\cvprfinalcopy 

\def\cvprPaperID{4987} 

\ifcvprfinal\fi
\begin{document}
\begin{spacing}{0.98}

\title{Monocular 3D Object Detection: An Extrinsic Parameter Free Approach} 

\author{
Yunsong Zhou \textsuperscript{\rm 1}~~ 
Yuan He \textsuperscript{\rm 2} \textsuperscript{\rm *}~~ 
Hongzi Zhu \textsuperscript{\rm 1} \textsuperscript{\rm *} ~~ 
Cheng Wang \textsuperscript{\rm 2}~~ 
Hongyang Li \textsuperscript{\rm 2}~~ 
Qinhong Jiang \textsuperscript{\rm 2,3} \\
\textsuperscript{\rm 1}Shanghai Jiao Tong University~~
\textsuperscript{\rm 2}SenseTime Research~~
\textsuperscript{\rm 3}Shanghai AI Laboratory
\\
{\{zhouyunsong,hongzi\}@sjtu.edu.cn}
{\{heyuan,wangcheng,lihongyang,jiangqinhong\}@senseauto.com}
}


\maketitle

\renewcommand{\thefootnote}{\fnsymbol{footnote}}
\footnotetext[1]{Co-corresponding authors}

\thispagestyle{empty}

\begin{abstract}
   Monocular 3D object detection is an important task in autonomous driving.
   It can be easily intractable where there exists ego-car pose change w.r.t. ground plane. This is common due to the slight fluctuation of road smoothness and slope. 
   Due to the lack of insight in industrial application, existing methods on open datasets \textbf{{neglect}} the camera pose information, which inevitably results in the detector being susceptible to camera extrinsic parameters. The perturbation of objects is very popular in most autonomous driving cases for industrial products.
   To this end, we propose a novel method to capture camera pose to formulate the detector free from extrinsic perturbation.
   Specifically, the proposed framework predicts camera extrinsic parameters by detecting vanishing point and horizon change. A converter is designed to rectify perturbative features in the latent space.
   By doing so, our 3D detector works independent of the extrinsic parameter variations and produces accurate results in realistic cases, e.g., potholed and uneven roads, where almost \textbf{all}\textit{} existing monocular detectors fail to handle.
   Experiments demonstrate our method yields the best performance compared with the other state-of-the-arts by a large margin on both KITTI 3D and nuScenes datasets. 
   
\end{abstract}

\section{Introduction}

3D object detection plays an important role in a variety of computer vision tasks, such as automated driving vehicles, autonomous drones, robotic manipulation, augmented reality applications, etc.
Most existing 3D detectors require accurate depth-of-field information. To acquire such resource, majority of the methods resort to the LiDAR pipeline \cite{chen_multiview_2016, qi_frustum_2017, shi_pointrcnn_2018, shin_roarnet_2018, liang_deep_2018, zhou_voxelnet_2017}, some to the radars solution \cite{major2019vehicle, yang2020radarnet, kimlow, lee2020deep} or others to the multi-camera framework \cite{chen_3d_2015, chen_3d_2018, li_stereo_2019, pham_robust_2017, qin_triangulation_2019, xu_multilevel_2018}.
In this paper, we address this problem in a monocular camera setting and curate it specifically for automated driving scenarios
With the difficulty in directly acquiring a depth of field information, monocular 3D detection (Mono3D) is an ill-posed and challenging task.
However, Mono3D approaches have the advantage of low cost, low power consumption, and easy-to-deployment in real-world applications.
Therefore, monocular 3D detection has received increasing attention
over the past few years \cite{brazil_m3d_rpn_2019, chen_monocular_2016, manhardt_roi_10d_2018, mousavian_3d_2017, qin_monogrnet_2018, simonelli_disentangling_2019}.

\begin{figure}[t]
\begin{center}
\includegraphics[width=\linewidth]{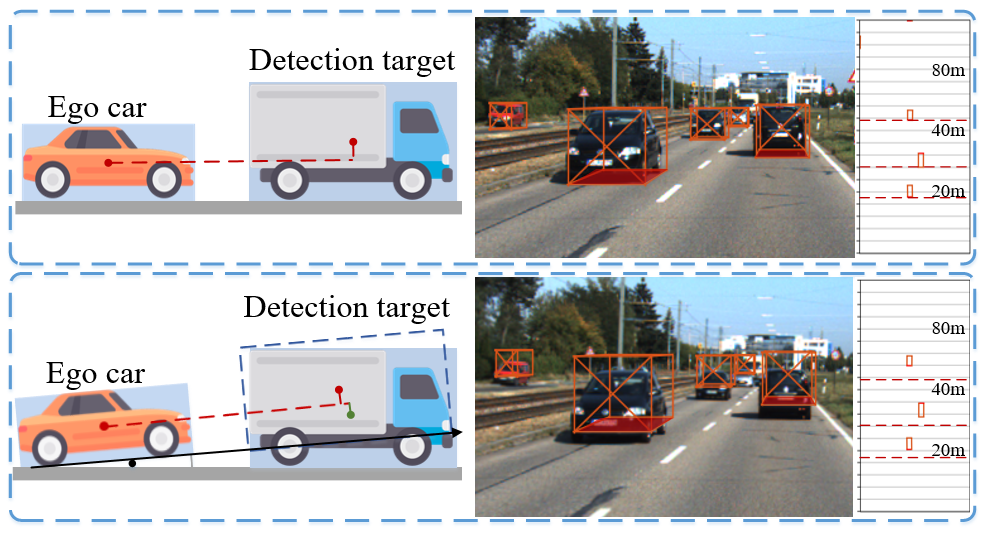}
\end{center}
   \caption{
   The \textbf{effect} of extrinsic parameter perturbations on 3D detection task.
   When the vehicle undergoes a slight pose change on an uneven road, the 3D detection results are less accurate (second row).
   This happens often in realistic applications and the detection offset can be viewed more evidently in the bird-eye's view.
   }
\label{fig:intro}
\vspace{-.5cm}
\end{figure}

Current Mono3D methods have achieved considerable high accuracy given a specifically fixed camera coordinate system.
However, in real scenarios, the unevenness (perturbation) of the road surface often causes the camera extrinsic parameters to be disturbed, which introduces a significant algorithmic challenge.
To the best of our knowledge, there are \textbf{\textit{no}} 3D detection datasets 
that takes into account the camera pose change under perturbation.

%

As shown in Figure \ref{fig:intro}, current datasets or detectors assume there is no perturbation, \textit{i.e.},  the extrinsic parameters are set to be constant. Therefore the accurate 3D results are obtained (top row). However, as depicted in the bottom perturbation case, the object information viewed by the camera deviates from the real object information. This makes the detection results unreliable by recovering a large offset in form of both 3D boxes and bird-eye's view.
Straightforward methods to address this problem are to design complementary branches or networks to improve the generalization ability, and yet this solution yields limited improvement \cite{chabot2017deep, xiang_subcategory-aware_2017,qin_monogrnet_2018, vianney2019refinedmpl, ding2020learning}. 
Some approaches utilize vehicle CAD models or keypoints to reconstruct vehicle geometry \cite{chabot2017deep, xiang_subcategory-aware_2017}, while others exploit existing networks to predict pixel-level or instance-level depth map by mimicking state-of-the-art (SOTA) LiDAR 3D detection methods, namely pseudo-LiDAR methods \cite{qin_monogrnet_2018, vianney2019refinedmpl, ding2020learning}.

Our work is inspired by the visual odometry methods that resolve camera pose change in adjacent frames from images \cite{godard2019digging,li2018undeepvo,ranjan2019competitive,wang2019recurrent,zhan2018unsupervised,zhou2017unsupervised}.
Note that this idea differentiates from those that focus \textit{solely} on detecting objects in the perturbation-prone {{camera coordinate system}} \cite{simonelli_disentangling_2019,brazil_m3d_rpn_2019,liu2020smoke,zhou_objects_2019,chen2020monopair,tang2020center3d,brazil2020kinematic}. These approaches focus on some less critical issues regarding to realistic industrial applications. For example, the modeling of occlusive objects \cite{chen2020monopair}, depth branches \cite{tang2020center3d}, kinematic motion information (object orientation) \cite{brazil2020kinematic}, \textit{etc}.
Moreover, it is similar to human behavior patterns that one can naturally adapt to changing road gradients and gradually deduce the accurate position of objects even on potholes. Formulating our network to encode such learning patterns is feasible on a biological basis.

In this paper, we propose to leverage the extrinsic parameter change implicitly in the image. Our key idea is to estimate camera pose change \textit{w.r.t.} the ground plane from images and optimize predicted 3D locations of objects guided by the camera extrinsic geometry constraint. We abbreviate the proposed framework as \textbf{MonoEF} (extrinsic parameter free detector).
Specifically, a novel detector is proposed to extract the vanishing point and horizon information from the image to estimate the camera extrinsic corresponding to the image.
The model is thus capable of capturing the extrinsic parameter perturbations to which the current image is subjected in the geometric space.
During inference, we transform latent feature space using extrinsic parameters as {seed} to remove the effect of extrinsic perturbations on features fed from the input image.
Note that the transformation network is learned in a supervised manner, which allows the image features to recover from camera perturbation.
By doing so, we impose our detector exclusive from the effects of the extrinsic parameter.
The resultant 3D locations are obtained via the extrinsic parameter-free predictor and projected back into the real-world coordinate system.

Experiments on both the KITTI 3D benchmark \cite{Geiger2012CVPR} and nuScenes dataset \cite{nuscenes2019} demonstrate that our method outperforms the SOTA methods by a large margin, especially for \textit{perturbative} examples with a distinguished improvement.
To sum up, the contributions of our paper are as follows:
\begin{itemize}
    \item We introduce a novel Mono3D detector by capturing the perturbative information of the extrinsic parameters from monocular images to make the detector free from extrinsic fluctuation.
    \item  We design a feature transformation network, using camera extrinsic parameters as seed, to recover the non-perturbative image information from the perturbative latent feature space.
    \item We propose an extrinsic module that complements the camera's pose in 3D object detection. Such a plug-and-play can be applied to existing detectors and pragmatic for industrial applications, \textit{e.g.}, autonomous driving scenarios.
\end{itemize}
The whole suite of the codebase will be released and the experimental results will be pushed to the public leaderboard.

\section{Related Work}


\vspace{2pt}\noindent\textbf{Monocular 3D Object Detection.}\hspace{5pt}
The Monocular camera is in lacks 3D information compared with multi-beam LiDAR or stereo cameras.
To overcome this difficulty and reconstruct the geometry and position of the object in world coordinates, most Mono3D methods can be roughly divided into three categories. In the first category \cite{chen_monocular_2016,murthy2017reconstructing,chabot2017deep}, auxiliary information is widely used like vehicle Computer-Aided Design (CAD) models or keypoints. By this means, extra labeling cost is inevitably required. In the second category \cite{ma_accurate_2019, manhardt_roi_10d_2018, wang_pseudo-lidar_2018, xu_multilevel_2018}, the prior knowledge like depth map by LiDAR point cloud, or disparity map by stereo cameras trained by external data is exploited. Usually, the inference time would increase significantly due to the prediction of these dense heat maps. 




\begin{figure*}[h]

\centering
\includegraphics[width=0.7\textwidth]{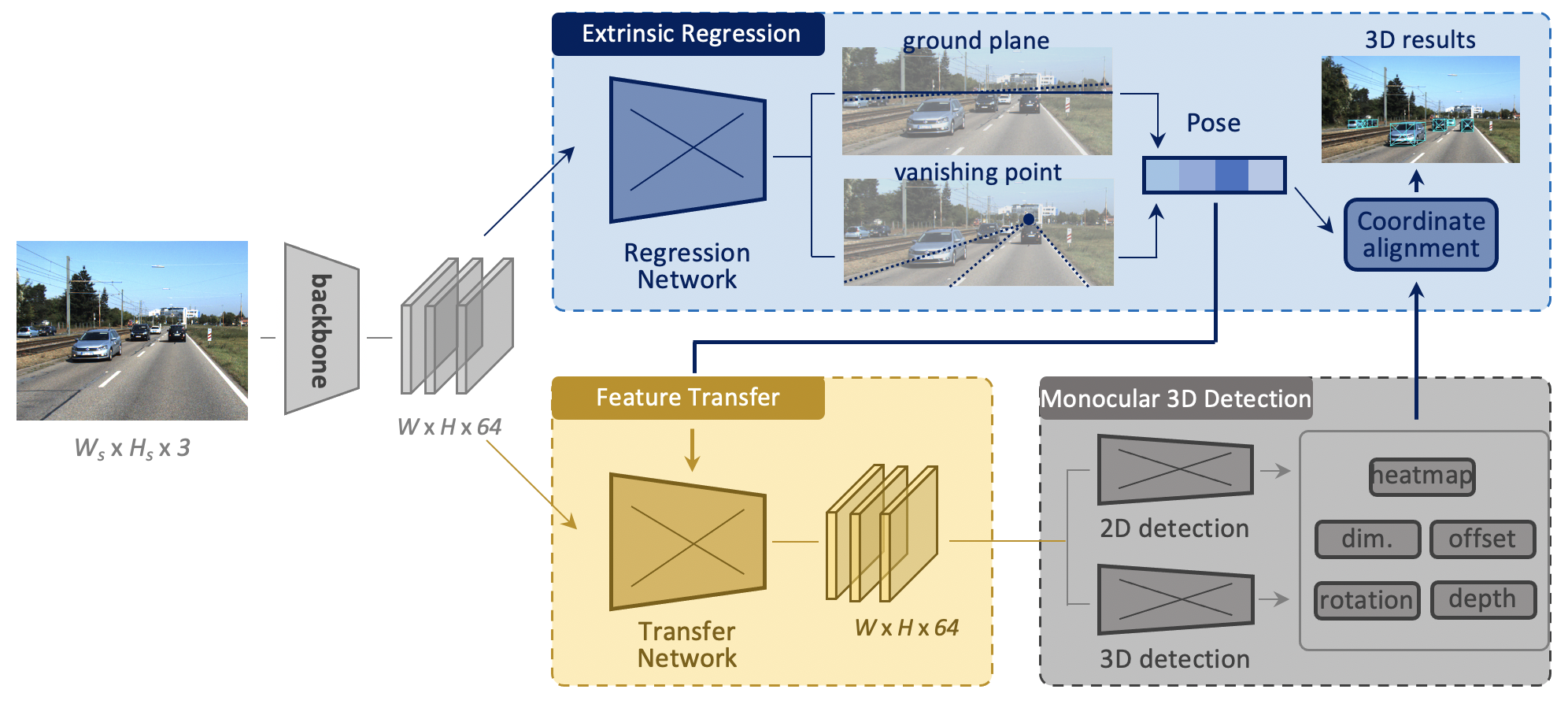}

\caption{\textbf{System overview.} The Extrinsic Regression module (blue block) predicts the ground plane as well as  vanishing point. 
   The pose information is thereby obtained and then fed into the Feature Transfer module (yellow block) as guidance for feature enhancement.
By doing so, the original features (in gray color) after the backbone are transferred to a rectified set of features (in yellow color), immune to the extrinsic parameter perturbation.
The Monocular 3D Detection module and coordinate alignment unit follow standard procedures \cite{liu2020smoke}.
	}
\label{fig:overview}
\end{figure*}

Unlike the aforementioned work, methods in the third category only make use of the RGB image as input and remove the dependencies on extra labeling or pre-trained networks by external data.
SMOKE \cite{liu2020smoke} predicts a 3D bounding box by combining a single keypoint estimation with regressed 3D variables based on CenterNet \cite{zhou_objects_2019}.
M3D-RPN \cite{brazil_m3d_rpn_2019} reformulates the monocular 3D detection problem as a standalone 3D region proposal network.
Current SOTA results for monocular 3D object detection are from MonoPair \cite{chen2020monopair}, Center3D \cite{tang2020center3d}, and Kinematic3D \cite{brazil2020kinematic}. Among them, MonoPair \cite{chen2020monopair}  improves the modeling of occlusive objects by considering the relationship of paired samples.
Center3D \cite{tang2020center3d} carefully designs two modules for better depth prediction called LID and DepJoint.
Kinematic3D \cite{brazil2020kinematic} proposes a novel method for monocular video-based 3D object detection which leverages kinematic motion to improve the precision of 3D localization.

However, all the object detectors mentioned above focus only on the information in the current camera coordinate system that ignores the effect of camera pose on detection.
These methods do not work well when the camera's pose receives a disturbance w.r.t. ground plane due to rough terrain or acceleration of ego vehicle.

%

\vspace{2pt}\noindent\textbf{Deep Monocular Odometry.}\hspace{5pt}
With the success of deep neural networks, end-to-end learning-based methods~\cite{wang2017deepvo,wang2018end,xue2018guided,xue2019beyond} have been proposed to tackle the visual odometry problem. Recently, some methods~\cite{bloesch2018codeslam,tang2018ba,teed2018deepv2d,ummenhofer2017demon,zhou2018deeptam} exploit CNNs to predict the scene depth and camera pose jointly, utilizing the geometric connection between the structure and the motion. This corresponds to learning Structure-from-Motion (SfM) in a supervised manner.
To mitigate the requirement of data annotations, self-supervised and un-supervised methods~\cite{godard2019digging,li2018undeepvo,ranjan2019competitive,wang2019recurrent,zhan2018unsupervised,zhou2017unsupervised} have been proposed to tackle the SfM task. 
CC \cite{ranjan2019competitive} addresses the unsupervised learning of several interconnected problems in low-level vision: single view depth prediction, camera motion estimation, optical flow, and segmentation of a video into the static scene and moving regions.
MonoDepth2 \cite{godard2019digging} proposes a set of improvements, which together result in both quantitatively and qualitatively improved depth maps compared to competing for self-supervised methods.
LTMVO \cite{zou2020learning} presents a self-supervised learning method for visual odometry with special consideration for consistency over longer sequences.

While these visual odometry methods are relatively good at detecting camera pose, they all rely on motion information on the time series, which will not be available in a typical Mono3D task based on single-frame images. Consequently, the lack of motion information in the time series prevents us from directly obtaining accurate camera pose information.
However, We can still use similar ideas to detect changes in the ground plane and vanishing point from the image compared to the reference frame, and thus indirectly infer changes in the camera extrinsic parameters.

\section{An Extrinsic Parameter Free Approach}
\subsection{Overview}\label{overview}

We adopt the one-stage anchor-free architecture as does in SMOKE \cite{liu2020smoke}.
Figure \ref{fig:overview} depicts an overview of our framework. 
It contains a backbone network, an extrinsic regression network, a feature transfer network, and several task-specific dense prediction branches.
The backbone takes a monocular image of size $(W_{s}\times H_{s}\times 3)$ as input and outputs a feature map of size $(W\times H\times 64)$ after down-sampling with an \textit{s}-factor.
The feature map is utilized for extrinsic parameter detection (top blue pipeline), and in parallel rectified by the transfer network based on extrinsic parameters (known as Pose as in the bottom yellow pipeline).
For 2D and 3D detection, we follow standard procedures in this domain. There exist seven output branches with each having size of $(W\times H\times m)$, where $m$ is the output channel of each branch.
The detection results need to be aligned by the predicted extrinsic parameters in order to get the final bounding box and position.

\begin{figure*}[h]

 \centering
		\includegraphics[width=0.7\textwidth]{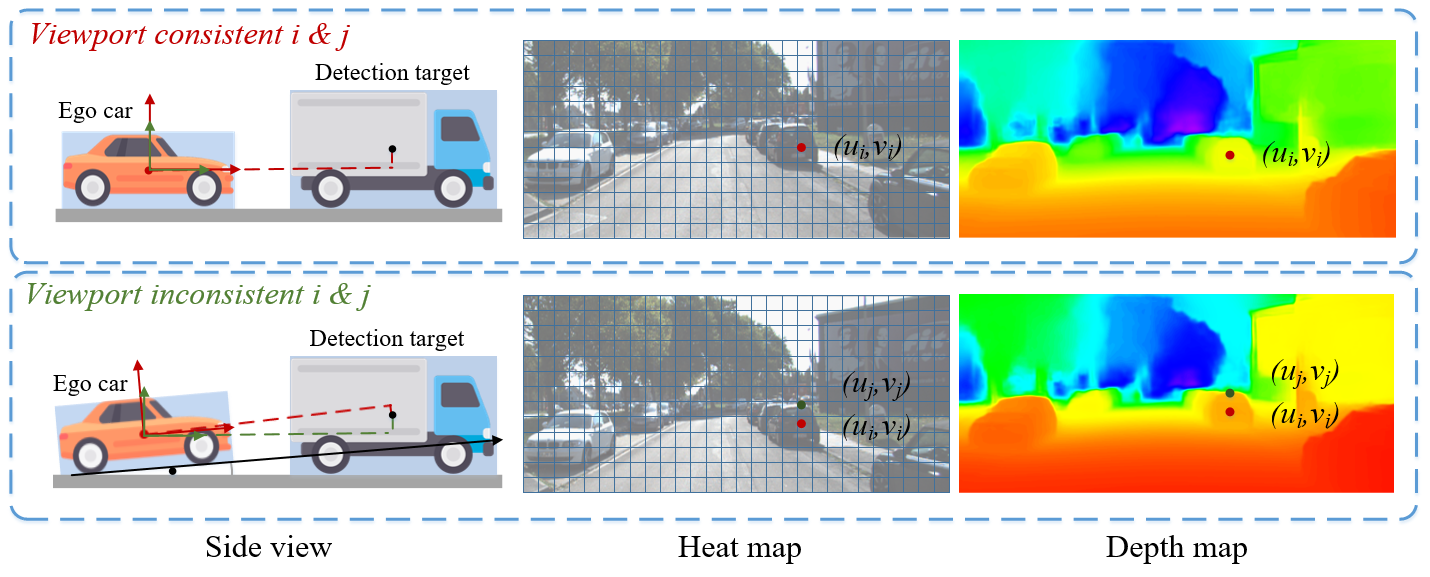}
	\caption{
	\textbf{Visualization} of the extrinsic perturbation.
   The pose of the ego vehicle varies due to the unevenness of road surfaces, which is quite common in realistic scenarios. It causes the camera's viewport $i$ to be inconsistent with ground viewport $j$.
   Therefore, the position of keypoints found from the heat map and depth map are shifted from $(u_{i}, v_{i})$ to $(u_{j}, v_{j})$ by extrinsic perturbation, leading to a confusion for the 3D prediction and thereby inaccurate results. 
	}
	\label{fig:car}
	\vspace{-0.4cm}
\end{figure*}

\subsection{Preliminary on Monocular Object Detection}

\textbf{The 2D object detection} follows the design of CenterNet \cite{zhou_objects_2019}.
A heatmap of size $(W\times H\times c)$ is used to enable keypoint localization$(u^g,v^g)$ and its classification.
The number of object categories $c$ equals three on KITTI3D benchmark and ten on nuScenes dataset.
The other two branches of size $(W\times H\times 2)$ are adopted to regress the dimensions of the 2D bounding box $(w^b,h^b)$ and the offset $(\delta^u,\delta^v)$ from the center of the bounding box $(u^b,v^b)$ to the keypoint $(u^g,v^g)$ correspondingly.

\textbf{The 3D object detection} focuses on the 3D information of an object in the local camera coordinate system instead of the global world coordinate system.
The object center in local camera coordinate system can be represented as homogeneous coordinates $\mathbf{c}^w=(x, y, z)$; its projection in the feature map is $\mathbf{c}^o=(u,v,1)$.
Similar to \cite{manhardt_roi_10d_2018, simonelli_disentangling_2019}, we predict the offset $(\Delta^u, \Delta^v)$ to the keypoint location $(u^g, v^g)$ and depth $z$ in two separate branches. 
Denote the coordinates in form of congruent concept, we have:
\begin{equation}
z \left[ \begin{matrix}   u  \   v  \  1  \end{matrix}  \right]^T =\mathbf{P}\cdot \left[
 \begin{matrix}   x  \  y \   z  \end{matrix}  \right]^T,
\end{equation}
where $\mathbf{P}$ is the projection conversion matrix between the world coordinate system and the image coordinate system. The projection matrix can be decomposed as:
\begin{equation}
    \mathbf{P}=\mathbf{K}\cdot \mathbf{T},
\end{equation}
where $\mathbf{K}$ is referred to as the constant camera intrinsic matrix and $\mathbf{T}$ 
as the inconstancy extrinsic matrix w.r.t ground plane. Naturally we have $\mathbf{c}^o=\frac{1}{z}\mathbf{P}\mathbf{c}^w$.
The depth $z$ and size $(w,h,l)$ are regressed according to \cite{eigen2014depth}.
As aforementioned in Section \ref{overview},
in these branches, the regression components are trained with the L1 loss. Similar to \cite{mousavian_3d_2017, zhou_objects_2019}, we represent the orientation using eight scalars, where the orientation branch is trained using the multi-bin loss \cite{mousavian20173d}.

\subsection{Theoretical Analysis}



Figure \ref{fig:car} depicts a concrete example of how an extrinsic perturbation can  significantly impose poor prediction onto heat map and depth map.

Given a specific local camera coordinate system called viewport $i$, it is normally assumed that the viewport $i$ is \textit{consistent} with ground plane coordinate system called viewport $j$, so do most of Mono3D datasets.
Suppose the 3D center of a selected object in viewport $i$ is $\mathbf{c}_i^w=(x_i,y_i,z_i)$, and the 3D center on the feature map is $\mathbf{c}_i^o=(u_i, v_i, 1)$, corresponding to the case as depicted in Figure \ref{fig:car}.
If there is an extrinsic perturbation from the ground plane variation, the identical relation between camera viewport $i$ and ground plane viewport $j$ would \textit{no longer} exist. We discriminate this process as \textit{perturbation}. The perturbation matrix $\mathbf{A}$ can be described as:
\begin{equation}
    \mathbf{A}=\left[ \begin{matrix}   \cos\theta_r & \sin\theta_r & 0  \\  \cos\theta_p\sin\theta_r & \cos\theta_r\cos\theta_p & \sin\theta_p \\  -\sin\theta_p\sin\theta_r & -\sin\theta_p\cos\theta_r & \cos\theta_p   \end{matrix}  \right] ,
    \label{A}
\end{equation}
where $\theta_p$ stands for pitch angle and $\theta_r$ for roll angle of ego vehicle w.r.t. ground plane respectively.
Now we are equipped with the extrinsic perturbation being introduced spatially, the center of the object $\mathbf{c}_i^w$ of camera viewport $i$ is transformed to a point $\mathbf{c}_j^w$ in the ground plane viewport $j$, where $\mathbf{c}_j^w=z_{j}\mathbf{P}_j^{-1}\mathbf{c}_j^o=\mathbf{A}\mathbf{c}_i^w$.
On the feature map, the keypoint of the object shifts correspondingly from $\mathbf{c}_i^o$ to $\mathbf{c}_j^o$.
The transfer relationship matrix $\mathbf{M}$ of keypoints on the feature map can be represented by:
\begin{equation}
    \mathbf{c}_j^o=\mathbf{M}\mathbf{c}^o_i=\frac{z_{i}}{z_{j}}\mathbf{P}_{j}\mathbf{A} \mathbf{P}_{i}^{-1}\mathbf{c}_i^o.
\end{equation}
This shift in image coordinates would cause confusion for the prediction of 3D position.


Given the example in Figure \ref{fig:car}, we use the changes on the depth hidden map to perform our analysis.
If the model can know the changes that occur in the ground plane coordinate system, such as LiDAR-based methods, it will assume that the target has changed in height. 
However, for the camera, the height and depth of the target will both affect its position on the image. 
The camera assumes that the target vehicle remains on the ground coordinate system at all times, so it incorrectly determines that the change in the target on the image is caused by the depth.
The offset of keypoints leads to large depth prediction errors.

\begin{figure}
    \centering
    \includegraphics[width=\linewidth]{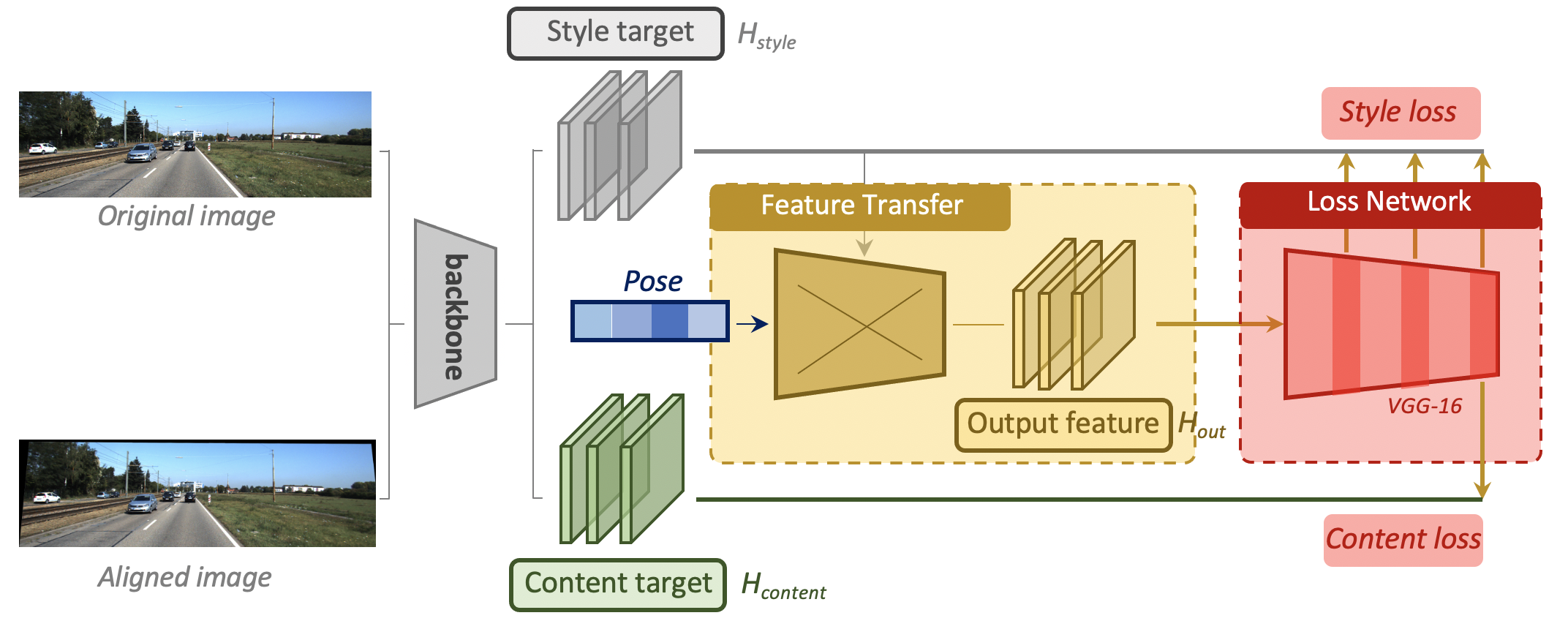}
    \caption{The training process of the transfer network $f^t$.
   The feature target is derived from a feature obtained by the backbone after a direct extrinsic parameter correction of the image.
   The pre-trained loss network $\Phi$ has two branches, one with the style target for high-dimensional losses computed on three layers and the other with the content target for low-dimensional losses computed only on the last layer.}
    \label{fig:style}
    \vspace{-0.5cm}
\end{figure}

These keypoint positions need to be rectified to compensate for offsets caused by the change of camera extrinsic parameter.
For training, $\mathbf{A}$ in Equation (\ref{A}) can be collected through the ground truth vehicle ego-pose information.
The image feature w.r.t. viewport $\mathbf{H}_j$ and labels $\mathbf{c}_j^w$ need to be first adjusted using matrices $\mathbf{M}^{-1}$ and $\mathbf{A}^{-1}$ separately in order to eliminate the effect of extrinsic perturbations $\mathbf{A}$.
The label obtained by 3D detector $f^{\theta_i}$ established at camera viewport $i$ can be recorded as $\hat{\mathbf{c}}_i^w$.
The L1 loss function under external parameter perturbation is changed to
\begin{equation}
\begin{split}
    \hat{\mathbf{c}}_i^w &= f^{\theta_i}(f^t(\mathbf{M}^{-1},\mathbf{H}_j)) = f^{\theta_i}(\mathbf{H}_i),\\
    L(\theta_j)&= \left \| \mathbf{A}^{-1}\mathbf{c}_j^w- \hat{\mathbf{c}}_i^w \right \| = \left \| \mathbf{c}_i^w -  \hat{\mathbf{c}}_i^w \right \|,
\end{split}
\end{equation}
where $f^t(\cdot,\cdot)$ is the transfer network on the feature implicit space which maps the change on camera extrinsic parameters to the feature map. 

During inference, we first estimate camera extrinsic parameters $\hat{\mathbf{A}}$ from input image $\mathbf{X}_j$ and recover the unperturbed feature hidden space $\hat{\mathbf{H}}_i$ from the perturbed feature hidden space $\mathbf{H}_j$ using $\hat{\mathbf{M}}$.
The predicted 3D center $\hat{\mathbf{c}}_j^w$ is derived from the 3D detector $f^{\theta_i}$ which is independent of varying camera extrinsic parameters $\hat{\mathbf{A}}$:
\begin{equation}
    \hat{\mathbf{c}}_j^w = \hat{\mathbf{A}}f^{\theta_i}(f^t(\hat{\mathbf{M}}^{-1},\hat{\mathbf{H}}_j)) = \hat{\mathbf{A}}f^{\theta_i}(\hat{\mathbf{H}}_i)).
\end{equation}


For camera extrinsic parameters $\mathbf{A}$ and $\mathbf{M}$, we propose the extrinsic regression network, which is introduced in Section \ref{Camera Extrinsic Parameters Regression}.
For feature transfer network $f^t$, the design methodology and training process is described in Section \ref{transfer net}.
These modules are utilized to detect extrinsic perturbations of the image in viewport $j$, and further adopt the extrinsic information to rectify the feature map.
In this way, the image features can be restored back to  camera viewport $i$, and the 3D detection model no longer receives the negative impact from extrinsic perturbations.

\subsection{Camera Extrinsic Parameters Regression}
\label{Camera Extrinsic Parameters Regression}

In addition to the regular regression task, we also introduce a module of extrinsic parameter regression in Mono3D branches, which is shown in Figure \ref{fig:overview}.

Owing to the fact that extrinsic parameters are too implicit for a model to regress, we choose to predict intuitive and explicit features from the image at first.
The horizon and vanishing point in the image are often used to help determine the vehicle's ego-pose information w.r.t ground plane in the deep visual odometry tasks. Specifically, the tilt of the horizon can indicate the change of roll angle, while the vertical movement of the vanishing point can indicate the change of pitch angle.

Following the SOTA odometry framework in \cite{chang2018deepvp}, we represent a regression task with L1 loss as:
\begin{equation}
\begin{split}
    [\hat{\mathbf{y}}_{\mathrm{gp}}, \hat{\mathbf{y}}_{\mathrm{vp}}]&=f^{\mathrm{vo}}(\mathbf{H}_j), \\
    L_{\mathrm{vo}}&=\left \| \mathbf{A}-\mathbf{g}(\hat{\mathbf{y}}_{\mathrm{gp}}, \hat{\mathbf{y}}_{\mathrm{vp}}) \right \|.
\end{split}
\end{equation}
Here, $f^{\mathrm{vo}}$ is the CNN architecture used for horizon and vanishing point detection, we follow \cite{krizhevsky2017imagenet} and make modifications to the filters for the fully connected layers.
$\hat{\mathbf{y}}_{\mathrm{gp}}$ and $\hat{\mathbf{y}}_{\mathrm{vp}}$ are the predicted ground plane and vanishing point results at viewport $j$.
The mapping function $\mathbf{g}:(\mathbb{R}^2, \mathbb{R}^2) \mapsto \mathbf{A}_{4\times 4}$ is a mathematical calculation function from the horizon and vanishing point to the camera extrinsic matrix.
The function $f^{\mathrm{vo}}$ ensures that the model can give sufficiently accurate information about the extrinsic parameters. Finally, the regression loss $L_{\mathrm{vo}}$ can be trained jointly with 2D and 3D detection branches.

\subsection{Feature Transfer by Extrinsic Parameters}
\label{transfer net}

To overcome the pose variation of ego vehicle w.r.t ground plane and improve 3D detection performance, we propose a transfer network applying camera extrinsic corrections on the feature latent layers.
Generally speaking, as shown in Figure \ref{fig:car}, the design intention of the transfer network is to rectify the perturbed feature space $\mathbf{H}_j$ in camera view $j$, so that the discrepancy between $\mathbf{H}_j$ and the unperturbed one $\mathbf{H}_i$ under camera view $i$ is as small as possible. 
For example, we fix the shift of keypoints caused by extrinsic parameter perturbations.
Suppose that in one image with unknown perturbation, the network predicts camera extrinsic parameters $\hat{\mathbf{A}}=\mathbf{g}(f^{\mathrm{vo}}(\mathbf{H}_j))$ based on the strategy in Section \ref{Camera Extrinsic Parameters Regression}.

After carefully analyzing the influence of perturbation on the image characteristics of low-dimensional features and high-dimensional features, we find out that their changing patterns are quite different. On the one hand, low-dimensional features like the position of corresponding edges and geometries are closely related to the camera's extrinsic parameters, specifically in terms of content information. On the other hand, high-dimensional features like the textures and illuminations remain unchanged, specifically in terms of style information.
Inspired by the image style transfer method \cite{johnson2016perceptual}, we propose a feature transfer module working on the feature latent space.


\begin{table*}[h]
\begin{center}
\small
\begin{tabular}{c|ccc|ccc|ccc|ccc|r}
\toprule
\multirow{2}{*}{Methods} & \multicolumn{3}{c|}{$AP_{3D}$}             & \multicolumn{3}{c|}{$AP_{BV}$}            & \multicolumn{3}{c|}{$AOS$}                & \multicolumn{3}{c|}{$AP_{2D}$}              & \multicolumn{1}{c}{\multirow{2}{*}{Time}} \\ \cline{2-13}
                         & E              & M              & H              & E              & M             & H              & E              & M              & H              & E              & M              & H              & \multicolumn{1}{c}{}                      \\ \midrule \midrule
M3D-RPN \cite{brazil_m3d_rpn_2019}                 & 14.76          & 9.71           & 7.42           & 21.02          & 13.67          & 10.23          & 88.38          & 82.81          & 67.08          & 89.04          & 85.08          & 69.26          & 0.16                                      \\
SMOKE \cite{liu2020smoke}                   & 14.03          & 9.76           & 7.84           & 20.83          & 14.49          & 12.75          & \textit{92.94} & \textit{87.02} & \textit{77.12} & 93.21          & 87.51          & 77.66          & \textbf{0.03}                             \\
MonoPair \cite{chen2020monopair}                 & 13.04          & 9.99           & 8.65           & 19.28          & 14.83          & 12.89          & 91.65          & 86.11          & 76.45          & \textbf{96.61} & \textbf{93.55} & \textbf{83.55} & 0.06                                      \\
PatchNet \cite{ma2020rethinking}                & 15.68          & 11.12          & \textit{10.17} & 22.97          & 16.86          & \textit{14.97} & -              & -              & -              & 93.82          & 90.87          & 79.62          & 0.40                                       \\
D4LCN \cite{ding2020learning}                   & 16.65          & 11.72          & 9.51           & 22.51          & 16.02          & 12.55          & 90.01          & 82.08          & 63.98          & 90.34          & 83.67          & 65.33          & 0.20                                       \\
Kinematic3D \cite{brazil2020kinematic}             & \textit{19.07} & \textit{12.72} & 9.17           & \textit{26.69} & \textit{17.52} & 13.1           & 58.33          & 45.5           & 34.81          & 89.67          & 71.73          & 54.97          & 0.12                                      \\ \midrule
\textbf{MonoEF}                   & \textbf{21.29} & \textbf{13.87} & \textbf{11.71} & \textbf{29.03} & \textbf{19.7}  & \textbf{17.26} & \textbf{96.19} & \textbf{90.65} & \textbf{82.95} & \textit{96.32} & \textit{90.88} & \textit{83.27} & \textbf{0.03}                             \\ \bottomrule
\end{tabular}
\end{center}
    \caption{$AP_{40}$ scores(\%) and runtime(s) on KITTI3D test set for car at 0.7 IoU threshold referred from the KITTI benchmark website. E, M and H represent \textit{Easy} , \textit{Moderate} and \textit{Hard} samples.
    Our model not only ranks first on the 3D evaluation metrics but also keeps the run time fairly low and comparable as a simple one-stage detection.
    Corner information might be cropped and padded by feature transferring and correction so that the performance of 2D detection is slightly affected.
    }
\label{KITTI bench}
\end{table*}


As shown in Figure \ref{fig:style}, this module is divided into two parts. One is the transfer network $f^t$, and the other is a pre-trained loss network $\Phi$ using \cite{simonyan2014very}.

\textbf{The transfer network.}
The input feature map $\mathbf{H}_{\mathrm{in}}$ for transfer network is provided by the previous backbone, which is equal to $\mathbf{H}_j$.
The predicted pose $\hat{\mathbf{M}}$ acts as a guidance information for transfer network, which provides structural information for feature maps in low dimensions.
The output of transfer network $\mathbf{H}_{\mathrm{out}}$ will be input into loss network with content target $\mathbf{H}_\mathrm{content}=f^b(\hat{\mathbf{M}}^{-1}\mathbf{X}_j)$ and style target $\mathbf{H}_\mathrm{style}=\mathbf{H}_j$ to calculate final features, where $\mathbf{X}_j$ stands for disturbed image input, and $f^b$ stands for backbone network.

\textbf{The loss network.}
The transfer network $f^t$ mainly considers content loss $l_{\mathrm{content}}$ and style loss $l_{\mathrm{style}}$.
Let $\phi_m$ be the activation of the $m$-th layer of the network $\Phi$ with the feature map of shape $(c_m\times h_m \times w_m)$.
The content feature reconstruction loss is the squared Euclidean distance between feature representations:
\begin{equation}
    l^{\phi,m}_{\mathrm{content}}(\mathbf{H}_{\mathrm{out}}, \mathbf{\mathbf{H}}_\mathrm{content})=\frac{\left\|\phi_m(\mathbf{H}_{\mathrm{out}})- \phi_m(\mathbf{H}_\mathrm{content}) \right\|_2^2}{c_mh_mw_m}.
\end{equation}
Following \cite{gatys2015texture}, we define the $Gram\ matrix\ G^{\phi}_m$ to be the $c_m\times c_m$ matrix whose elements are given by:
\begin{equation}
    G^{\phi}_m(\mathbf{H})_{c,c'}=\frac{\sum^{h_m}_{h=1}\sum^{w_m}_{w=1}\phi_m(\mathbf{H})_{h,w,c}\phi_m(\mathbf{H})_{h,w,c'}}{c_mh_mw_m}.
\end{equation}
The Gram matrix can be computed by reshaping $\phi_m(\mathbf{H})$ into a matrix $\psi$, then $G^{\phi}_m(\mathbf{H})=\psi \psi^T/c_mh_mw_m$.
The style reconstruction loss is then the squared Frobenius norm of the difference between the Gram matrices of the output and target feature maps:
\begin{equation}
    \l^{\phi,m}_{\mathrm{style}}(\mathbf{H}_{\mathrm{in}},\mathbf{H}_\mathrm{style})=\left \| G^{\phi}_m(\mathbf{H}_{\mathrm{in}})-G^{\phi}_m(\mathbf{H}_\mathrm{style}) \right \|_F^2.
\end{equation}
The $l_{\mathrm{content}}$ penalizes the output feature map when it deviates in content from the target and $l_{\mathrm{style}}$ penalizes differences in style.
The joint total loss is defined as:
\begin{equation}
    L_{\mathrm{total}}=\gamma_1l_{\mathrm{content}}+\gamma_2l_{\mathrm{style}},
\end{equation}
where $\gamma_1$ and $\gamma_2$ are hyper-parameters for tuning content loss and style loss.

\begin{table*}[]
\small
\begin{center}
\begin{tabular}{c|c|ccc|ccc|ccc}
\toprule
\multirow{2}{*}{Methods} & \multirow{2}{*}{Test Data} & \multicolumn{3}{c|}{$AP_{3D}$}              & \multicolumn{3}{c|}{$AP_{BV}$}             & \multicolumn{3}{c}{$AP_{2D}$}            \\ \cline{3-11} 
                         &                           & E              & M              & H              & E              & M              & H              & E             & M             & H              \\ \midrule \midrule
\multirow{3}{*}{M3D-RPN \cite{brazil_m3d_rpn_2019}} & original                  & 64.91          & 50.53          & 41.73          & 69.28          & 53.64          & 45.12          & 91.88         & 93.11         & 77.24          \\
                         & disturbed                 & 39.72          & 31.08          & 25.73          & 48.37          & 38.55          & 32.22          & 92.20         & 93.14         & 77.13          \\ \cline{2-11} 
                         & decrease                  & -25.19         & -19.45         & -16.00            & -20.91         & -15.09         & -12.9          & \textbf{0.32} & \textbf{0.03} & \textit{-0.11}          \\ \midrule
\multirow{3}{*}{SMOKE \cite{liu2020smoke}}   & original                  & 77.89          & 72.80          & 65.37          & 83.30          & 82.92          & 75.76          & 99.50         & 99.05         & 90.55          \\
                         & disturbed                 & 42.58          & 35.09          & 30.74          & 53.01          & 44.15          & 39.41          & 98.66         & 98.12         & 89.84          \\ \cline{2-11} 
                         & decrease                  & -35.31         & -37.71         & -34.63         & -30.29         & -38.77         & -36.35         & -0.85         & -0.92         & -0.71          \\ \midrule
\multirow{3}{*}{D4LCN \cite{ding2020learning}}   & original                  & 61.54          & 45.60          & 37.77         & 68.32          & 51.68          & 39.31          & 97.35         & 89.1         & 71.51          \\
                         & disturbed                 & 41.77          & 29.22          & 25.78          & 59.9          & 43.45          & 36.06          & 85.38         & 76.64         & 60.03          \\ \cline{2-11} 
                         & decrease                  & \textit{-19.77}         & \textit{-16.38}         & \textit{-11.99}         & \textit{-8.42}         & -8.23         & \textit{-3.25}         & -11.97         & -12.46         & -11.48          \\ \midrule
\multirow{3}{*}{Kinematic3D \cite{brazil2020kinematic}}   & original                  & 55.45          & 39.47          & 31.29         & 61.72          & 44.65          & 34.58          & 98.61         & 86.3         & 71.39          \\
                         & disturbed                 & 27.30          & 16.95          & 13.79          & 47.78          & 36.70          & 29.24          & 90.84         & 55.13         & 44.80          \\ \cline{2-11} 
                         & decrease                  & -28.15         & -22.52         & -17.50         & -13.94         & \textit{-7.95}         & -5.34         & -7.77         & -31.17         & -26.59          \\ \midrule
\multirow{3}{*}{\textbf{MonoEF}}  & original                  & 77.55          & 72.83          & 72.01          & 82.33          & 82.80          & 75.61          & 99.56         & 99.19         & 90.62          \\
                         & disturbed                 & 76.87          & 70.86          & 63.86          & 81.64          & 74.76          & 73.92          & 99.65         & 99.15         & 90.62          \\ \cline{2-11} 
                         & decrease                  & \textbf{-0.68} & \textbf{-1.97} & \textbf{-8.16} & \textbf{-0.68} & \textbf{-8.04} & \textbf{-1.70} & \textit{0.09}          & \textit{-0.04}         & \textbf{-0.01} \\ \bottomrule
\end{tabular}
\end{center}
    \caption{$AP_{40}$ scores(\%) on KITTI3D validation set for car at 0.5 IoU threshold before and after camera extrinsic disturbance. 
    The lower the decreased value, the better the performance.
    The original target coordinates are transformed according to the pitch and roll angle set by the artificial extrinsic perturbation.
    The input image is also processed using the projection transformation according to these angles.
    }
\label{disturbed}
\end{table*}

\section{Experimental Results}

\subsection{Implementation Setup}

We conduct experiments on the KITTI3D object detection dataset, KITTI odometry dataset and nuScenes dataset. The KITTI3D dataset does not collect camera extrinsic information, which means its $\mathbf{T}$ matrix is an identity matrix. We can only find vehicle ego-pose information from the KITTI odometry and nuScenes datasets.

For the evaluation and ablation study, we show experimental results from two different setups. \textbf{Baseline} is derived from SMOKE \cite{liu2020smoke} with an additional output branch for camera extrinsic parameters.
\textbf{MonoEF} is the final proposed method integrating seven prediction branches, camera extrinsic parameter regression branch, and camera extrinsic amendment network.

For the rest of the detailed dataset statistics, training and inference structure, learning rules, evaluation metrics, \textit{etc.}, please refer to the supplementary.

\subsection{Quantitative and Qualitative Results}

\begin{figure*}[t]
\begin{center}
\includegraphics[width=0.98\linewidth]{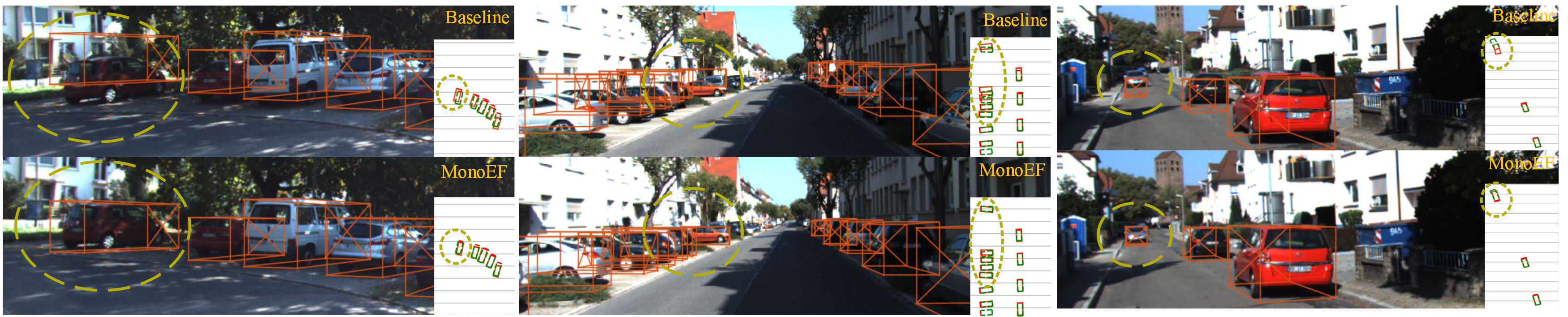}

\end{center}
   \caption{
   Qualitative results on KITTI odometry dataset. 
   The prediction 3D bounding boxes of SMOKE (the one above) and our model (the one below) are shown under camera extrinsic perturbation in the images.
   Green boxes and orange boxes in bird view mean ground truth and predictions of cars.
   A more pronounced difference in the predictions appears where the dashed line is circled. It can be seen from the figure that our model is effective against the perturbation of the external participants, especially for depth prediction.
   }
\label{fig:result_fig}
\end{figure*}

We first show the performance of our proposed MonoEF on KITTI3D object detection benchmark\footnote{http://www.cvlibs.net/datasets/kitti/eval\_object.php?obj\_benchmark=3d} for car. Comparison results with other state-of-the-art (SOTA) monocular 3D detectors including M3D-RPN \cite{brazil_m3d_rpn_2019},  SMOKE \cite{liu2020smoke}, MonoPair \cite{chen2020monopair}, PathNet \cite{ma2020rethinking}, D4LCN \cite{ding2020learning} and Kinematic3D \cite{brazil2020kinematic} are shown in Table \ref{KITTI bench}.
$AP_{2D}$ and $AOS$ are metrics for 2D object detection and orientation estimations following the benchmark.
We achieve the highest score for all kinds of samples and rank in first place among those 3D monocular object detectors on other metrics, regardless our model is only comparable or a bit worse than SOTA detector MonoPair \cite{chen2020monopair} on $AP_{2D}$.
Our method outperforms Kinematic3D for a large margin in $AP_{3D}$ and $AP_{BV}$, especially for $Hard$ samples.
The comparison of results fully proves the effectiveness of the proposed camera extrinsic amendment for images with unknown perturbations.

\begin{table}[]
\small
\begin{center}
\begin{tabular}{c|c|ccc}
\toprule
Class                             & Methods & ATE $\downarrow$        & ASE $\downarrow$      & AOE $\downarrow$     \\ \midrule \midrule
\multirow{2}{*}{car}              & baseline & 0.73          & 0.16          & 0.09          \\
                                  & MonoEF   & \textbf{0.56} & \textbf{0.15} & \textbf{0.09} \\ \midrule
\multirow{2}{*}{pedestrian}       & baseline & 0.85          & 0.32          & 1.48          \\
                                  & MonoEF   & \textbf{0.71} & \textbf{0.31} & \textbf{0.99} \\ \midrule
\multirow{2}{*}{motorcycle}       & baseline & 0.84          & 0.23          & 0.86          \\
                                  & MonoEF   & \textbf{0.70} & \textbf{0.23} & \textbf{0.79} \\ \midrule
\multirow{2}{*}{\textbf{overall}} & baseline & 0.87          & 0.57          & 0.75          \\
                                  & MonoEF   & \textbf{0.77} & \textbf{0.37} & \textbf{0.65} \\ \bottomrule
\end{tabular}
\end{center}
\caption{Evaluation errors on the nuScenes test dataset.
Our errors are lower on the three representative categories selected.
In overall classes, our error is much lower than baseline.
}
\label{nuscene}
\end{table}

Table \ref{disturbed} shows the performance on KITTI3D validation set for the car with and without camera extrinsic perturbation.
Since the KITTI3D dataset is initially without perturbation information of the camera pose, we simulate the camera extrinsic parameter perturbation in the real world using an artificially set Gaussian function ($pitch, roll \sim N(0, 1)$).
We evaluate the related values of SOTA monocular detectors through their published detection models.
It can be noticed that the detection performance of all models is degraded more or less after the addition of the extrinsic perturbation.
The other models are quite sensitive to extrinsic perturbations, with very severe performance degradation, while our model only has a slight performance drop.
This demonstrates the effectiveness of our model in handling camera extrinsic perturbations.
Figure \ref{fig:result_fig} shows the qualitative results on KITTI odometry dataset.
In this dataset we can get the vehicle pose information, so we know the real-world extrinsic parameter perturbations to which the vehicle is subjected.
The parts drawn with dashed lines indicate that our model has good performance against perturbations, especially in depth estimation.

\begin{table}[]
\small
\begin{center}
\begin{tabular}{c|c|c}
\toprule
Category                         & Methods       & Angular Error $\downarrow$ \\ \midrule \midrule
\multirow{3}{*}{Multiple frames} & CC \cite{ranjan2019competitive}    & 0.0320        \\ 
                                 & MonoDepth2 \cite{godard2019digging}    & 0.0312        \\
                                 & LTMVO \cite{zou2020learning}        & \textbf{0.0142}        \\ \midrule
Single frame                  & MonoEF        & \textit{0.0287}        \\ \bottomrule
\end{tabular}
\end{center}
\caption{Angular Error(deg/m) on KITTI Odometry validation sequence 08.
Methods designed specifically for the odometry task use information from consecutive frames to detect pose, and we have achieved comparable detection accuracy by doing the detection only on a single frame.
}
\label{angle error}
\end{table}

Since there is no open source code the more challenging nuScenes dataset by time, we only evaluate our model on it, which is shown in Table \ref{nuscene}.
From this dataset, we can get ego car pose information.
The table shows smoke of the more representative categories in the dataset, and we can see that our model's prediction errors on these categories have decreased compared to the baseline.
Across all categories, our model reduced the overall ATE and ASE quite a lot.
This demonstrates the enhancement of our model for the 3D detection task on the nuScenes dataset.

\subsection{Ablation Study}

We conduct several ablation studies for different evaluation items and data settings. 
We only show results from $Moderate$ samples here.

\textbf{Time expense analysis.}
Other Mono3D models may require some additional operations to assist the prediction during inference, such as generating pseudo-lidar \cite{brazil2020kinematic}, generating pairs \cite{chen2020monopair}, \textit{etc}.
Compared to these methods, MonoEF is based on the SMOKE \cite{liu2020smoke} with modified extrinsic parameters and only needs to go through a backbone network during the inference process. 
We can see from Figure \ref{KITTI bench} that our method also has a great advantage in time expense.

\textbf{Camera pose detection.}
For camera extrinsic parameters regression study, we evaluate the angular errors of the MonoEF on the KITTI odometry verification sequence 08, comparing with SOTA monocular visual odometry methods including CC \cite{ranjan2019competitive}, MonoDepth2 \cite{godard2019digging} and LTMVO \cite{zou2020learning}.
The evaluation results shown in Table \ref{angle error} indicate that although our model is not specifically designed to implement visual odometry functionality, it is also possible to predict accurate camera poses and achieve SOTA performance on the KITTI odometry dataset.
This ensures the accuracy of the camera extrinsic parameters regression.

\begin{table}[]
\small
\begin{center}
\begin{tabular}{c|c|ccc}
\toprule
Data                   & Methods  & $AP_{3D}$ & $AP_{BV}$ & $AP_{2D}$ \\ \midrule \midrule
\multirow{2}{*}{Single Seq.}    & baseline & 34.98  & 43.29 & \textbf{80.51}   \\
                       & MonoEF   & \textbf{41.78}  & \textbf{52.78} & 79.33    \\ \midrule
\multirow{2}{*}{Multiple Seq.} & baseline  & 23.46  & 26.51 & 75.41  \\
                       & MonoEF   & \textbf{26.06}  & \textbf{32.43} & \textbf{80.21}  \\ \bottomrule
\end{tabular}
\end{center}
\caption{$AP_{40}$ scores(\%) evaluated on KITTI Odometry sequecnce 00 (trained on single sequence 00) and sequecnce 08 (trained on sequence 00-07 \& 09-10) for car.}
\label{single sequence}
\end{table}

\begin{table}[]
\small
\begin{center}
\begin{tabular}{l|ccc}
\toprule

Methods              & $AP_{3D}$ & $AP_{BV}$  & $AP_{2D}$ \\ \midrule \midrule
M3D-RPN \cite{brazil_m3d_rpn_2019}              & 36.13           & 42.88         & 67.49             \\
M3D-RPN \cite{brazil_m3d_rpn_2019} + E.F.        & \textbf{41.36}  & \textbf{43.41} & \textbf{67.57}   \\ \midrule
Kinematic3D \cite{brazil2020kinematic}          & 41.44           & 43.51         & 65.70             \\
Kinematic3D \cite{brazil2020kinematic} + E.F.    & \textbf{48.92}  & \textbf{51.39} & \textbf{66.92}   \\ \midrule
SMOKE \cite{liu2020smoke}               & 34.98           & 43.29       & \textbf{80.51}       \\
SMOKE \cite{liu2020smoke} + E.F. (MonoEF)     & \textbf{41.78}  & \textbf{52.78}   & 79.33      \\ \bottomrule

\end{tabular}
\end{center}
\caption{$AP_{40}$ scores(\%) evaluated on KITTI Odometry sequecnce 00 for SOTA methods, including M3D-RPN \cite{brazil_m3d_rpn_2019},  Kinematic3D \cite{liu2020smoke} and SMOKE \cite{brazil2020kinematic}. +E.F. indicates that we apply the transfer network to feature maps by extrinsic regression network to the original method.}
\label{several methods}
\end{table}

\textbf{Camera extrinsic amendment.}
In terms of the camera extrinsic amendment study, we perform performance comparison experiments on sequences of the KITTI odometry dataset shown in Table \ref{single sequence}.
Because the odometry dataset does not contain a 3D detection label, we used the point cloud detection model 3DSSD \cite{yang20203dssd} to formulate the ground truth.
For the detection task training on the single sequence and multi sequences, our model shows a substantial improvement on performance with the camera extrinsic amendment compared to the baseline.
The improvement is more pronounced on a single sequence since the initial frame of different sequences in KITTI odometry dataset can not assure a consistent camera pose w.r.t. ground plane, which would confuse the extrinsic regression network.
We apply our MonoEF to other SOTA detection models \cite{brazil_m3d_rpn_2019, liu2020smoke, brazil2020kinematic} and achieve similar significant improvements, which is shown in Table \ref{several methods}.

\section{Conclusion}

We propose a novel method for monocular 3D object detection with two camera-extrinsic-aware modules, namely the extrinsic regression net and the feature transfer net.
By capturing the camera pose change from image \textit{w.r.t} ground plane and performing a corresponding amendment for the naturally ill-posed Mono3D detection, our method is robust against camera extrinsic perturbation and helps  model predict much more accurate depth results.
Our model achieves the state-of-the-art performance on KITTI3D object detection benchmark using a monocular camera and proves its efficiency on KITTI odometry and nuScenes dataset.

\section*{Acknowledgements}
This research was supported in part by National Key R\&D Program of China (Grant No. 2018YFC1900700) and National Natural Science Foundation of China (Grants No. 61772340).


\end{spacing}
\end{document}